# A Knowledge-Guided Cross-Modal Feature Fusion Model for Local Traffic Demand Prediction

Lingyu Zhang, Pengfei Xu, Guobin Wu, Jian Liang, Ruiyang Dong, Yunhai Wang, Xuan Song

***Abstract*—Traffic demand prediction plays a critical role in intelligent transportation systems. Existing traffic prediction models primarily rely on temporal traffic data, with limited efforts incorporating human knowledge and experience for urban traffic demand forecasting. However, in real-world scenarios, traffic knowledge and experience derived from human daily life significantly influence precise traffic prediction. Such knowledge and experiences can guide the model in uncovering latent patterns within traffic data, thereby enhancing the accuracy and robustness of predictions. To this end, this paper proposes integrating structured temporal traffic data with textual data representing human knowledge and experience, resulting in a novel knowledge-guided cross-modal feature representation learning (KGCM) model for traffic demand prediction. Based on regional transportation characteristics, we construct a prior knowledge dataset using a large language model combined with manual authoring and revision, covering both regional and global knowledge and experiences. The KGCM model then learns multimodal data features through designed local and global adaptive graph networks, as well as a cross-modal feature fusion mechanism. A proposed reasoning-based dynamic update strategy enables dynamic optimization of the graph model's parameters, achieving optimal performance. Experiments on multiple traffic datasets demonstrate that our model accurately predicts future traffic demand and outperforms existing state-of-the-art (SOTA) models.**

This work was partially supported by the grants of Jilin Provincial International Cooperation Key Laboratory for Super Smart City and Jilin Provincial Key Laboratory of Intelligent Policing.

Lingyu Zhang is with School of Computer Science and Technology, Shandong University, Qingdao, China and Research Institute of Trustworthy Autonomous Systems, Southern University of Science and Technology (SUSTech), Shenzhen, China. (e-mail: zhanglingyu@mail.sdu.edu.cn)

Pengfei Xu is with School of Information Sciences and Technology, Northwest University, Xi'an, China. (pfxu@nwu.edu.cn)

Guobin Wu and Jian Liang are with Didi Chuxing, Beijing, China. (liangjian@didiglobal.com, wuguobin@didiglobal.com)

Ruiyang Dong is with Department of Computer Science and Engineering, Southern University of Science and Technology, Shenzhen, China. (dongry2022@mail.sustech.edu.cn)

Yunhai Wang is with Renmin University of China Beijing, China. (wang.yh@ruc.edu.cn)

Xuan Song is with School of Artificial Intelligence, Jilin University, Changchun, China and Research Institute of Trustworthy Autonomous Systems, Southern University of Science and Technology (SUSTech), Shenzhen, China. (songxuan@jlu.edu.cn)

## I. INTRODUCTION

SPATIOTEMPORAL information prediction is a key technology in intelligent transportation systems (ITS), playing a central role in advancing smart urban transportation [1]. Current mainstream prediction methods rely on historical spatiotemporal data and provide decision-making support for diverse users. These methods offer real-time services such as accurate travel time prediction and pick-up point prediction for individual travelers, and also generate macro-level indicators such as traffic flow and travel demand forecasts for government and enterprise decision-makers [2]. However, increasingly complex urban traffic systems and dynamically changing external environments create significant challenges for traffic prediction technologies.

The core scientific challenge in intelligent traffic prediction lies in constructing effective spatiotemporal dependency modeling mechanisms that can simultaneously capture the temporal evolution patterns and spatial correlation characteristics from traffic data. Specifically, the traffic flow in a particular region is influenced by its own historical temporal patterns and the dynamic changes in geographically adjacent areas. To overcome the limitations of traditional approaches that emphasize temporal modeling while neglecting spatial dependencies, related spatiotemporal prediction models [3,4] have achieved significant breakthroughs by jointly modeling spatial and temporal correlations. Notably, with the expansion of urban areas and the development of public transportation networks, the influence of long-range spatial dependencies and long-term temporal patterns on traffic prediction has become increasingly prominent. On the one hand, non-adjacent regions may interact dynamically through the transmission mechanisms of transportation networks. On the other hand, there may exist latent semantic correlations between areas that are far apart from each other. To address these challenges, researchers have proposed multi-level modeling strategies. In the temporal dimension, the models based on temporal convolutional networks (TCNs) [7,8], recurrent neural networks (RNNs) [9 – 11], and their attention-enhanced variants [12–15] deeply explore the periodic and trend features of traffic flow. In the spatial dimension, graph convolutional networks (GCNs) [16 – 19] are employed to model road network topologies, and stacked multi-layer convolutions [5] are used to capture long-range spatial correlations. Additionally, hybrid modeling approaches that integrate domain knowledge [6] have shown unique advantages in specific scenarios. These technological advancements have significantly improved prediction accuracy and accelerated the practical deployment of intelligent transportation systems [20–22].

Recent works on spatiotemporal prediction have demonstrated that constructing efficient spatiotemporal feature extraction mechanisms is a critical pathway to achieving accurate forecasting. However, existing frameworks still exhibit significant limitations when applied to real traffic scenarios. For the current models, the features extracted from single-modal traffic data are difficult to provide more valuable information. However, existing models ignore human knowledge and experience, which is a very important auxiliary information for traffic prediction. From cognitive science, travelers accumulate spatiotemporal knowledge through habitual routes, forming an important prior knowledge. This reservoir of experience interacts with real-time traffic data to jointly influence travel decisions. Specifically, a user's traffic prediction behavior essentially manifests as a dynamic cognitive integration mechanism that not only involves the interpretation of real-time sensor data but also incorporates historical experience (such as memories of congestion on specific routes during commuting hours) and socialized information (such as traffic incident alerts disseminated through social media). This composite decision-making process significantly surpasses the traditional prediction paradigm that relies solely on sequential traffic data.

Existing traffic prediction models exhibit significant shortcomings in feature representation and knowledge integration, which lead to the prediction results of existing models struggling to achieve the expected predictive effect. Fig. 1 illustrates the comparison results of traditional traffic prediction models and knowledge-guided traffic prediction models, respectively, on the New York City dataset. As illustrated in Fig. 1(a), traditional spatiotemporal neural networks often suffer from prediction deviations in complex scenarios, such as shifts in traffic patterns during holidays or the propagation of impacts from unexpected incidents, primarily due to a lack of prior knowledge.

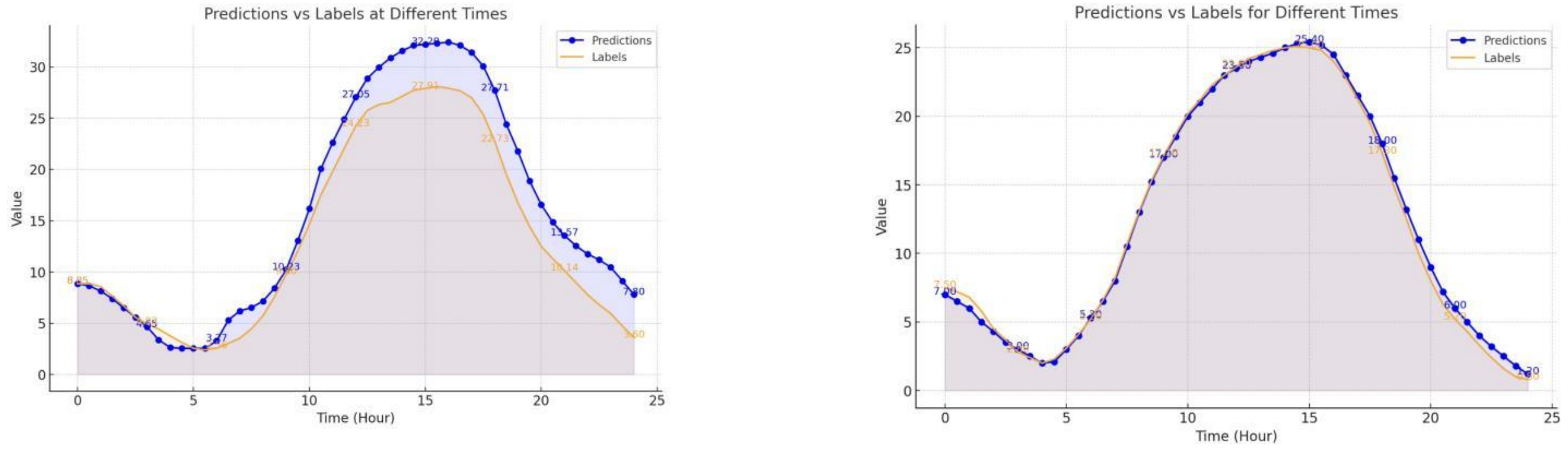

(a)Traditional traffic prediction models without prior knowledge guidance

(b) Knowledge-guided traffic prediction model based on multimodal feature fusion

**Fig. 1.** Comparison results of traditional traffic prediction models and knowledge-guided traffic prediction model on the taxi public dataset from New York City (NYC) Taxi & Limousine Commission, TLC (October and November, 2024)

Although knowledge-guided traffic prediction has significant theoretical advantages and practical value, there are still several critical technical challenges. For example, it is hard to construct a quantifiable feature representation for traffic knowledge, and also difficult to design effective knowledge injection mechanisms to prevent model overfitting. To solve these problems, we propose a novel knowledge fusion framework between traffic knowledge and spatiotemporal traffic data through a collaborative learning mechanism, and achieving deep integration of data-driven features and domain knowledge. Experimental results demonstrate that the proposed model improves peak-period prediction accuracy by up to 23.7%, and significantly enhances prediction stability under abnormal road network conditions compared to traditional models. As shown in Fig.1(b), multimodal feature fusion techniques are designed to jointly encode both structured domain knowledge and unstructured experiential knowledge, and the prediction results match the labels much better.

Incorporating experiential textual information into the feature extraction of temporal traffic data can further enrich the feature representations for traffic prediction. However, this approach also presents several challenging issues, such as building a diverse and comprehensive dataset of traffic knowledge, accurately and effectively aligning semantic features from multimodal data, and achieving collaborative feature learning between global/regional traffic data and corresponding experiential knowledge. To address these issues, this paper proposes a knowledge-guided cross-modal feature fusion (KGCM) model for local traffic demand prediction. KGCM integrates traffic knowledge and experience into the temporal feature extraction, and employs the strategies of cross-attention and two-stage feature learning to effectively extract representative features, thereby improving the model's overall performance and robustness are improved.

By modeling and analyzing multi-source heterogeneous data (including historical orders, real-time traffic conditions, weather information, and large-scale events), the proposed traffic demand prediction model enables more accurate and fine-grained forecasting of future demand across different regions. This innovative approach effectively addresses a longstanding challenge of intelligent transportation: the inability of traditional reactive models to manage supply-demand imbalances caused by sudden demand fluctuations, and significantly enhances the operational efficiency and service quality of transportation platforms, which further offers substantial value in optimizing resource allocation, improving user satisfaction, strengthening market competitiveness, and promoting the coordinated development of smart urban mobility systems.

The main contributions of this paper are summarized as follows:

1. We proposes an innovative traffic demand prediction model by fusing the textual and temporal traffic data. By leveraging textual data derived from traffic knowledge and experiences, the model is guided to focus on fine-grained features within the temporal traffic data. Multimodal feature fusion enhances the accuracy and robustness of traffic demand forecasting.

2. We design a regional-level multimodal fusion module based on a guided cross-attention mechanism and a global-level multimodal fusion module based on a relational matrix. These two modules fully exploit fine-grained and global-level experiential knowledge and are integrated into a two-stage training process. Through dynamic adjustment of both local and global model parameters, optimal network performance is achieved.

3. We conduct extensive experiments on multiple traffic datasets, including the taxi public dataset from NYC Taxi & Limousine Commission, TLC (October and November, 2024), a private dataset in Chengdu form DiDi ( November, 2016), and a private Bike-sharing dataset form The City of New York's bicycling data (October and November, 2024). The final results demonstrate that our proposed model significantly outperforms the state-of-the-art traffic demand prediction model. In specific time periods or regions, the introduction of experiential knowledge significantly improves prediction accuracy. Specifically, on the NYC taxi dataset, our model achieves relative improvements of 0.58%, 3.30%, and 12.66% in MAPE, MAE, and RMSE metrics, respectively. On the Chengdu dataset, it achieves relative improvements of 4.37%, 13.16%, and 20.52% in MAPE, MAE, and RMSE metrics, respectively. On the Bike-sharing dataset, it achieves relative improvements of 1.84%, 6.98%, and 3.72% in MAPE, MAE, and RMSE metrics, respectively.

## II. RELATED WORKS

### *2.1 Traditional Traffic Prediction Methods Based on Temporal Neural Networks*

In recent years, traffic prediction has received widespread attention from both academia and industry. Traditional traffic prediction methods based on temporal neural networks, such as Recurrent Neural Networks (RNN), Long Short-Term Memory (LSTM) networks, and Gated Recurrent Units (GRU), have shown significant advantages in temporal feature learning. However, due to the dynamic nature of traffic system, long-term traffic prediction remains challenges. To address these problems, He et al. proposed a Spatiotemporal Convolutional Neural Network (STCNN) based on Convolutional LSTM to learn spatiotemporal correlations from historical traffic data for long-term prediction [23]. Subsequently, existing segment-based prediction models often neglect regional spatial dependencies, then He et al. proposed a Spatiotemporal Neural Network (STNN) further [24]. STNN uses an encoder-decoder architecture, and incorporates region-based spatial dependencies and external factors to improve accuracy. To capture both spatial and temporal dependencies simultaneously, Zhao et al. proposed a Temporal Graph Convolutional Network (T-GCN), which combines Graph Convolutional Networks (GCN) and GRU to capture spatial dependencies via complex topological structures and temporal dependencies[25]. Then, some relevant works explored spatiotemporal relationships using 2D CNNs and LSTM. Chen et al. proposed an end-to-end Multi-Gated Spatiotemporal Convolutional Neural Network (MGSTC) for urban-scale flow prediction. MGSTC explores dependencies through multiple gated branches and dynamically integrates spatiotemporal features with external factors [26]. For subway and access network prediction, Wang et al. developed a Transformer-based spatiotemporal method using multi-head attention to capture temporal correlations and graph convolution layers for spatial correlations [27]. Furthermore, existing methods are designed based on the assumption of invariant spatiotemporal correlations. Then, Li et al. proposed AST3DRNet, a spatiotemporal 3D residual network with an attention mechanism, to directly predict the congestion levels of urban road networks by effectively modeling spatiotemporal information [28].

Existing models can effectively capture nonlinear and dynamic variations in traffic data, but these methods exhibit limitations in modeling long-range dependencies and handling multiscale, multisource heterogeneous data. Consequently, recent research works have shifted towards more sophisticated spatiotemporal hybrid modeling methods by incorporating graph neural networks, attention mechanisms, and other advanced techniques.

### *2.2 Traditional Traffic Prediction Methods Based on Graph Convolutional Networks*

Traffic flow analysis, prediction, and management underpin modern smart city development, and deep neural networks and massive traffic data facilitate understanding complex traffic network patterns. Traditional Graph Convolutional Networks (GCN) have emerged as a vital tool in traffic prediction due to their strong modeling capability for graph-structured data.

Although there have some related works have addressed future traffic flow prediction, they also have limitations in modeling spatiotemporal dependencies. For example, Wang et al. proposed a novel spatiotemporal graph neural network by introducing a learnable position attention mechanism to aggregate adjacent road information. This network used a sequence modeling component to capture traffic flow dynamics, and leveraged both local and global temporal dependencies [29]. Xie et al. proposed a Self-Attention-based Spatiotemporal Graph Neural Network (SAST-GNN) to efficiently capture highly nonlinear dependencies and integrate spatiotemporal features. By incorporating a self-attention mechanism, SAST-GNN accurately extracts temporal and spatial features, so as to enhance further by channel and residual blocks for multi-perspective fusion [30]. By considering both the static mixed urban traffic network structure and dynamic spatiotemporal relationships, Peng et al. proposed a spatiotemporal associated dynamic graph neural network framework [31]. They designed a Dynamic Graph Convolutional Recurrent Neural Network (Dynamic-GRCNN) to learn spatiotemporal feature representations of urban traffic network topology and transportation hubs. Spatiotemporal graph-based deep learning methods show excellent performance, but their separate modeling of spatial interactions and temporal dependencies often leads to inefficient post-fusion, causing prediction delays and biases. To solve this problem, Wang et al. proposed a Traffic Gated Graph Neural Network (Traffic-GGNN) for real-time spatiotemporal fusion. Traffic-GGNN tries to capture location-based spatial interactions via bidirectional message-passing and dynamically feature fusion[32]. To addresse the limitations of fixed graph structures and improper spatiotemporal dependency extraction, Sun et al. proposed a Spatiotemporal Graph Neural Network based on Adaptive Neighbor Selection (STGNN-ANS). This method employs a neighbor selection mechanism to filter inappropriate neighbors and generate new graph structures. To capture spatiotemporal dependencies, STGNN-ANS utilizes Bidirectional LSTM (BiLSTM) and a self-attention enhanced GCN to achieve high accuracy in short-distance and long-distance scenarios [33]. Additionally, Geng et al. proposed a Spatiotemporal Gated Attention Transformer model (STGAFormer) based on graph neural networks to enhance long-term prediction and handle emergencies. STGAFormer incorporates a distance-based spatial self-attention module with a threshold screening mechanism to selectively identify key features from nearby and distant areas [34].

Although GCNs have the advantages of extracting spatial dependencies, challenges persist, including the limitations of static adjacency matrices and insufficient modeling of dynamic temporal dependencies. Thus, recent research works increasingly explore new methods combining temporal models, graph attention mechanisms, and adaptive graph learning to surpass traditional GCN performance bottlenecks.

### *2.3 Traffic Prediction Based on Multimodal Information Fusion*

With the advancement of smart cities and intelligent transportation systems, a single data source is no longer sufficient to meet the requirements for high-accuracy predictions. Consequently, numerous related studies have increasingly focused on traffic prediction based on multimodal information fusion, attempting to integrate various heterogeneous data sources (such as traffic flow data, road topology, weather, holidays, geographic information, and events) to comprehensively model the dynamic characteristics of traffic systems.

Considering that most studies emphasize the spatiotemporal correlations of traffic flow, which are considered unstable under different conditions, and false correlations may exist in observational data, Zhao et al. analyzed the physical concepts affecting the generation of multimodal traffic flow from the perspective of observation generation principles, and further proposed a Causal Conditional Hidden Markov Model (CCHMM) for multimodal traffic flow prediction. This model uses a mutual-supervised training method with prior and posterior knowledge to enhance model identifiability, effectively decoupling causal representations of interested concepts and identifying causal relationships to accurately predict multimodal traffic flow [35]. Meanwhile, to utilize the short-term non-stationarity and spatiotemporal correlations in traffic flow, Yin et al. proposed a spatiotemporal hybrid prediction model ST-VGBiGRU based on improved Variational Mode Decomposition (VMD), Graph Attention Networks (GAT) and Bidirectional Gated Recurrent Units (BiGRU)[36]. Existing research works often ignores the comprehensive analysis of spatiotemporal distributions and integration of multimodal representations, then Zhou et al. proposed a large-scale spatiotemporal multimodal fusion framework based on location queries. This framework not only effectively demonstrates its ability to integrate multimodal data in spatiotemporal hyperspace but has also been successfully applied to real-world large-scale maps[37]. With the development of large-scale models, traffic prediction technology has made significant progress. However, various events happen in cities, such as sports competitions, exhibitions, concerts, etc., which significantly impact traffic patterns in surrounding areas and may cause advanced prediction models to fail in such situations. Therefore, Han et al. proposed a simple and effective multimodal event traffic prediction model that uses pre-trained text encoders and traffic encoders to extract embedding vectors, which are then fused for prediction. This model aims to extend the applicability of traffic prediction technology, focusing on modeling the impact of events on traffic patterns [38]. Given that the correlation between predicted traffic flow and recent historical traffic flow has been continuously strengthening, most existing traffic flow prediction models typically learn spatiotemporal patterns from single-scale historical traffic data. However, such short-term traffic data does not contain continuous and dynamic long-term spatiotemporal patterns. Therefore, to comprehensively learn diverse patterns and achieve a satisfactory balance between effectiveness and efficiency has become a challenge. Then, Lv et al. proposed a multimodal urban traffic flow prediction model based on multi-scale time-series imaging (MM-TSI). By designing a multi-scale time-series imaging data processing mechanism, MM-TSI efficiently learns both short-term and long-term spatiotemporal patterns, and can effectively learn more comprehensive and diverse spatiotemporal patterns while maintaining certain efficiency [39]. Furthermore, Large language models (LLMs) have also demonstrated outstanding performance in traffic flow prediction tasks. Although LLMs can capture spatiotemporal dependencies in traffic flow prediction, they often neglect the cross-relationships between spatiotemporal embeddings. To address this problem, Xu et al. proposed a spatiotemporal fusion model for traffic flow prediction (GPT4TFP), which introduces a multi-head cross-attention-based spatiotemporal fusion strategy to capture the cross-relationships between spatiotemporal embeddings [40].

Although related research on traffic flow prediction based on multimodal information fusion has made certain progress, it still needs to face the challenges of low data homogeneity, complex relationships between modalities, and limited model generalization ability. Therefore, how to make breakthroughs in multimodal feature fusion, cross-modal alignment, and joint optimization has become a key research topic.

## III. Traffic Demand Prediction Based on Knowledge-Guided Cross-Modal Feature Fusion Learning

To achieve accurate traffic demand prediction in complex urban traffic environments and provide decision support for urban traffic management and taxi service optimization, this paper proposes a multi-stage temporal modeling framework based on knowledge-guided cross-modal feature fusion learning. This framework fully integrates structured temporal data, regional text description information, and cross-regional common semantic cues. By introducing adaptive graph structure learning, cross-modal fusion mechanisms, and dynamic feature enhancement, as shown in Fig.2. We achieve accurate traffic demand prediction through two stages. In the first stage, a local prior knowledge guidance mechanism is used for multimodal data fusion, fully exploiting the complementary features between structured temporal data and local textual information. To effectively model the dynamic dependencies and temporal evolution trends between features, a graph structure dynamic optimization mechanism is then introduced to capture the hidden dependencies between features, providing more expressive intermediate representations for downstream tasks. In the second stage, a global prior knowledge guidance mechanism further integrates local features with regional common information, enhancing the model's synergy and generalization. Finally, the fused features, weighted by the adaptive relationship matrix, are input into the structure-aware Informer to obtain the final traffic demand prediction results.

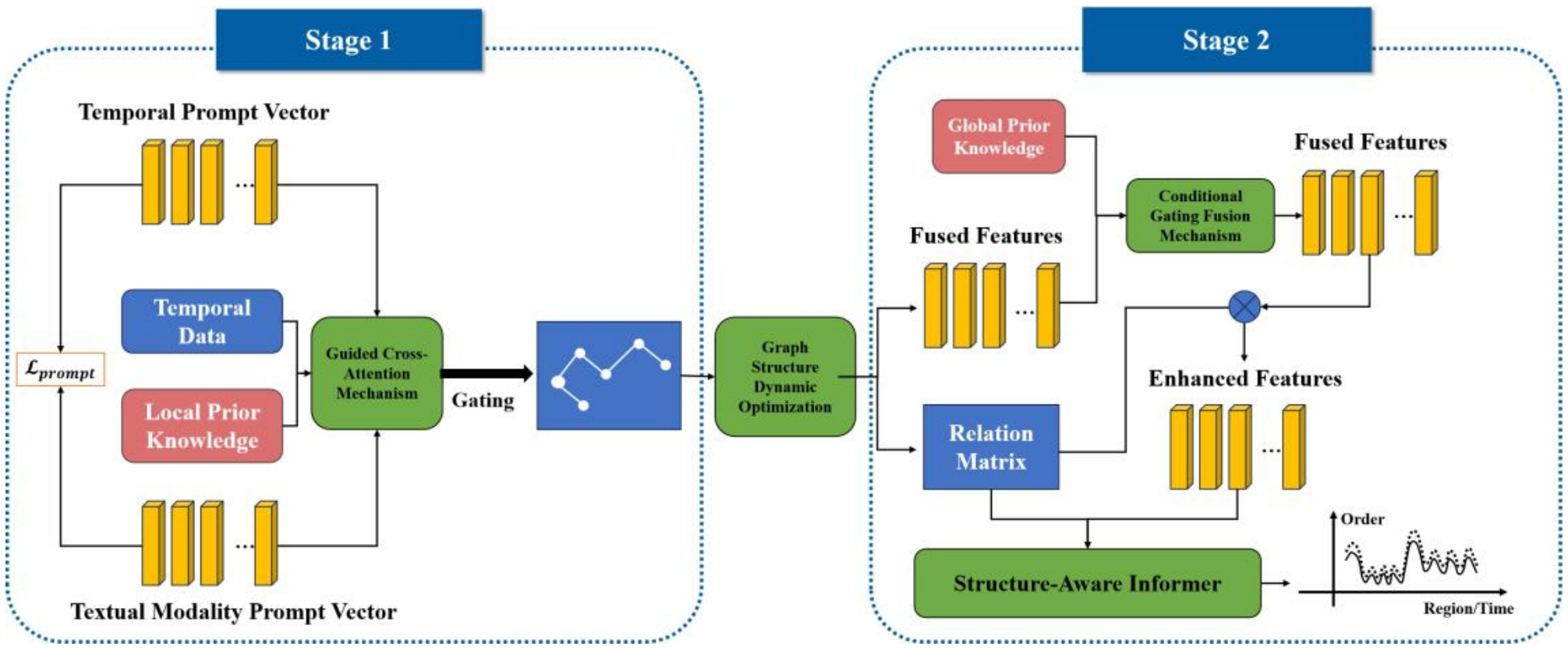

**Fig 2.** Multi-stage traffic demand prediction framework based on knowledge-guided cross-modal feature fusion learning.

### *3.1 The Traffic Prior Knowledge*

Traditional temporal prediction tasks typically rely on pattern recognition within historical data. Introducing textual prior knowledge into this process helps the model identify some of the nonlinear or implicit relationships in the historical data, enhancing its generalization ability. In this paper, we obtained the traffic data by aggregating the New York City Taxi Public Dataset, Chengdu Private Dataset, and Bike Private Dataset from October to November. To acquire the corresponding prior knowledge of the traffic data, we first manually give some description for a small portion of taxi orders with experiential knowledge, and we can obtain the initial textual data. Then, a large language model (ChatGPT) is used to generate similar textual descriptions for the remaining traffic data. Finally, we manually corrected the generated textual descriptions based on actual conditions to obtain accurate textual information.

Compared to traditional methods that rely on numerical data for traffic demand prediction, incorporating textual prior knowledge offers the following advantages: First, textual prior knowledge provides richer information on data variations, and can help the model understand the underlying factors and improve the prediction accuracy. Second, in cases where historical numerical data is insufficient or missing, textual descriptions can fill this gap by providing additional information, enabling the model to make effective predictions even in data-scarce environments. Furthermore, text can intuitively express periodic features, such as "weekday peak hours" or "holidays," which helps the model recognize cyclical changes in order volume and enhance prediction accuracy. Lastly, by combining numerical data with both local and global textual data in multimodal learning, the model can extract features from different data sources, promote information integration, and make more comprehensive predictions.

### *3.2 Local Prior Knowledge Guidance Module*

To fully explore the complementary features between structured temporal data and the corresponding textual information over time, we designs a multimodal fusion module based on local prompt optimization (Local Prompt Optimization, LPO). This module not only accurately aligns information across modalities but also guides the model to focus on more effective temporal or semantic signals through prompt vectors, thereby enhancing the accuracy and generalization ability of the order volume prediction model.

Let the structured input at time $t$ be represented as the vector $\mathbf{x}_t \in \mathbb{R}^d$, which includes features such as order volume, average passenger count, and average distance. A feedforward network is used for embedding:

$$\boldsymbol{h}_t^s = ReLU\left(\boldsymbol{W}_s \boldsymbol{x}_t + \boldsymbol{b}_s\right), \boldsymbol{h}_t^s \in \mathbb{R}^d \quad (1)$$

For each time point, let the corresponding textual description be $T_t$. We use BERT's Tokenizer to convert the text into a format acceptable to the model and add special tokens (such as [CLS] and [SEP], which are the modules in BERT). Then, the processed text is input into the BERT model to obtain its corresponding contextual embedding representation.

$$\boldsymbol{h}_t^t = BERT_{CLS}\left(T_t\right), \boldsymbol{h}_t^t \in \mathbb{R}^d \quad (2)$$

The representations of the two modalities, $\boldsymbol{h}_t^s$ and $\boldsymbol{h}_t^t$, are aligned in the same dimensional space.

To capture the implicit correspondence between the structured and textual modalities, we introduce an improved guided cross-attention mechanism that integrates prompt vectors to enhance the adaptiveness of attention weights and semantic guidance capability. We introduce independent learnable prompt vectors for each modality, namely, the structured modality prompt $\boldsymbol{p}_s \in \mathbb{R}^d$ and the textual modality prompt $\boldsymbol{p}_t \in \mathbb{R}^d$. By controlling the attention construction process with the prompts, we make it more semantically targeted. We use the structured modality data as the query and the local textual modality data as the key-value pairs to compute attention.

$$\begin{aligned} \boldsymbol{Q}_t &= \boldsymbol{h}_t^s \boldsymbol{W}_Q + \boldsymbol{p}_s \\ \boldsymbol{K}_t &= \boldsymbol{h}_t^t \boldsymbol{W}_K + \boldsymbol{p}_t \quad (3) \\ \boldsymbol{V}_t &= \boldsymbol{h}_t^t \boldsymbol{W}_V \end{aligned}$$

The fused representation obtained by the structured modality after perceiving its semantics through the textual modality is as follows:

$$\boldsymbol{z}_t = SoftMax\left(\frac{\boldsymbol{Q}_t \cdot \boldsymbol{K}_t^\top}{\sqrt{d}}\right)\boldsymbol{V}_t \quad (4)$$

The gated fusion mechanism automatically adjusts the importance allocation between the structured features and the textual modality. The structured raw representation and the cross-attention fusion results are then integrated.

$$\boldsymbol{g}_t = \sigma\left(\boldsymbol{W}_g\left[\boldsymbol{h}_t^s; \boldsymbol{z}_t\right]\right) \quad (5)$$

$$\boldsymbol{h}_t^{fusion} = \boldsymbol{g}_t \odot \boldsymbol{h}_t^s + \left(1 - \boldsymbol{g}_t\right) \odot \boldsymbol{z}_t \quad (6)$$

Here, $\odot$ represents element-wise multiplication, $\sigma$ is the sigmoid activation function, $[* ; *]$ denotes vector concatenation, $\boldsymbol{W}_g$ is the learnable parameter, and $\boldsymbol{g}_t$ controls the fusion ratio between structured features and textual information. To encourage semantic consistency between the two modality prompt vectors, we design a mutual prompt loss, which constrains the semantic consistency of the prompts through a regularization term, enhancing the modality fusion effect.

$$\mathcal{L}_{prompt} = \| \boldsymbol{p}_s - \boldsymbol{p}_t \|_2^2 \quad (7)$$

This loss function ensures that the model maintains semantic alignment between the two modalities during the fusion process, reducing issues of information inconsistency. Ultimately, the fused representation at each time point, $\boldsymbol{h}_t^{fusion} \in \mathbb{R}^d$ is obtained.

### *3.3 Dynamic Graph Structure Optimization*

To effectively model the dynamic dependencies and temporal evolution trends between multimodal features, this study proposes a dynamic graph structure optimization strategy (Dynamic Graph Structure Optimization, DGSO). The core of this module lies in the adaptive construction of structural relationships between features and hierarchical modeling of temporal dependencies, aiming to achieve deep interaction modeling of multiple features within local regions, thereby providing more expressive intermediate representations for downstream tasks.

Traditional graph structure learning methods typically use global static graphs, which are unable to capture the dynamic changes in feature dependencies over time. However, in complex traffic prediction or urban computing scenarios, the interactions between multimodal features (such as holidays, textual descriptions, passenger counts, etc.) vary greatly over different time periods. To adapt to this dynamic nature, this paper does not adopt a static feature connection structure but instead builds an adaptive feature relationship graph based on learning to capture the implicit dependencies between features at any given time. After multimodal fusion, let there be $d$ feature dimensions (i.e., nodes in the graph), and each node has a learnable low-dimensional embedding representation. At each layer $l$ and each time $t$, the corresponding feature state is $\mathbf{H}_t^{(l,fusion)} \in \mathbb{R}^{d\times n}$, which is mapped into the following two relational representation matrices:

$$\mathbf{Q}_t^{(l)} = \mathbf{H}_t^{(l,fusion)}\mathbf{W}_Q \quad (8)$$

$$\mathbf{K}_t^{(l)} = \mathbf{H}_t^{(l,fusion)}\mathbf{W}_K \quad (9)$$

Where $\mathbf{W}_Q, \mathbf{W}_K \in \mathbb{R}^{N\times N'}$ are learnable parameters. Therefore, we define the relationship matrix between features as:

$$\boldsymbol{A}_t^{(l)} = SoftMax\left(ReLU\left(\mathbf{Q}_t^{(l)} \cdot \left(\mathbf{K}_t^{(l)}\right)^{\top}\right)\right) \quad (10)$$

$$\boldsymbol{A}_t^{(l)} = \boldsymbol{\lambda} \cdot \boldsymbol{A}_{t-1}^{(l)} + (\mathbf{1} - \boldsymbol{\lambda}) \cdot \boldsymbol{A}_{\boldsymbol{t}}^{(l)} \quad (11)$$

Where $\boldsymbol{A}_t^{(l)} \in \mathbb{R}^{d\times d}$ represents the relationship matrix between features, indicating the degree of dependency of each feature dimension on other features. This matrix can capture, for example, the impact intensity of "holidays" on "order volume."

After obtaining the feature relationship graph, we introduce a structure-aware graph convolution mechanism for feature representation learning at each time point. This mechanism is a dynamic graph convolution operation, allowing each feature to dynamically integrate information from all other features based on its relevance. Thus, the model's ability to express key features is enhanced. Let the output of the $l-1$ layer graph convolution be $\boldsymbol{H}^{(l-1,fusion)} \in \mathbb{R}^{d\times n}$, and we define the update rule for the $l$-th layer features as follows:

$$\boldsymbol{H}^{(l,fusion)} = \sigma\left(\boldsymbol{A}^{(l)} \cdot \boldsymbol{H}^{(l-1,fusion)} \cdot \boldsymbol{W}^{(l)}\right) \quad (12)$$

Where $\boldsymbol{H}^{(l-1,fusion)}$ represents the node representations from the previous layer's output. $\boldsymbol{W}^{(l)}$ represents the trainable weights of the current layer. $\sigma$ is the nonlinear activation function. To improve the stability of deep networks and alleviate the vanishing gradient problem, we introduce residual connections and layer normalization(LN) operations in each layer.

$$\boldsymbol{H}_{out}^{(l,fusion)} = \mathrm{LN}\left(\boldsymbol{H}^{(l,fusion)} + \boldsymbol{H}^{(l-1,fusion)}\right) \quad ($$

*3.4 Global Prior Knowledge Guidance Module*

This module aims to further enhance the expressive power of temporal features by further integrating local features with regional common semantic cues, thereby improving the synergy and generalization of the temporal prediction model. Specifically, we use the fused feature representation $\boldsymbol{H}_{out}^{fusion}$ obtained in the first stage, along with the feature-level adaptive relationship matrix $\boldsymbol{A} \in \mathbb{R}^{d\times d}$ learned through the graph network, and combine them with global semantic cues to construct input features suitable for long-sequence prediction tasks.

Considering that different regions often share similar external influencing factors during certain periods (such as holidays, weather, events, etc.), we propose a Regional Common Prompt Guidance Mechanism (RCPG). This mechanism introduces text description information $T_{global}$ shared across regions, which depicts the common features and behavioral patterns of all sub-regions during the current period in natural language. We use BERT's Tokenizer to convert the text into a format acceptable to the model and add special tokens (such as [CLS] and [SEP]) where necessary. Then, the processed text is input into the BERT model to obtain the regional common prompt vector.

$$\boldsymbol{p}_{global} = BERT_{CLS}\left(T_{global}\right) \in \mathbb{R}^d \quad (14)$$

Next, for the fused representation $\boldsymbol{H}_t^{fusion}$ at each time point $t$, we design a conditional gated fusion mechanism to dynamically learn the fusion ratio between it and the global prompt vector. First, we compute the fusion gating factor $\boldsymbol{g}_t'$:

$$\boldsymbol{g}_t' = \sigma\left(\boldsymbol{W}_1 \cdot \left[\boldsymbol{H}_t^{fusion}; \boldsymbol{p}_{global}\right] + \boldsymbol{b}_1\right) \quad (15)$$

$$\tilde{\boldsymbol{h}}_t = \boldsymbol{g}_t' \odot \boldsymbol{H}_t^{fusion} + \left(1 - \boldsymbol{g}_t'\right) \odot \boldsymbol{p}_{global} \quad (16)$$

Where $\odot$ denotes element-wise multiplication, $\sigma$ is the sigmoid activation function, and $\boldsymbol{W}_1$ and $\boldsymbol{b}_1$ are learnable parameters. This mechanism can be viewed as an information injection-based prompt fusion strategy, which controls the injection of common prompt information through gating, thereby providing the prediction model with stronger regional common perception capabilities. This operation ensures that each region's representation at the current time step not only contains its information but also integrates shared information from relevant sub-regions, enabling each time step's representation to have dual expressive power of local regional information and global collaborative semantics, enhancing the model's collaborative prediction ability.

Next, we further focus on the structural dependencies between the features within the fused representation. Although the aforementioned fusion operation has achieved effective integration of structured data, local textual descriptions, and regional common prompt information through cross-modal interaction learning, enhancing semantic expressiveness, this operation assumes that the dimensions of the feature vectors are independently and identically distributed, neglecting the semantic coupling and collaborative dependencies between different feature dimensions. For example, the order volume feature may highly depend on factors such as holidays, weather, price, and some regional static attributes. Meanwhile, the representation obtained by merging structured features and text often contains redundant, irrelevant, or potentially conflicting information from multiple sources, which can interfere with learning key temporal features. The adaptive feature relationship matrix provides a data-driven mechanism to weaken the influence of irrelevant or redundant features. Through structural weighting, the model can recombine feature representations to obtain feature vectors with higher semantic consistency and discriminative power, helping to enhance the model's robustness and generalization ability. Therefore, we propose an Adaptive Correlation Matrix-based Feature Weighting strategy (ACMFW), which uses the feature-level adaptive matrix $\boldsymbol{A} \in \mathbb{R}^{d\times d}$ obtained in the first stage to further weight the fused representation $\tilde{\boldsymbol{h}}_t$:

$$\widehat{\boldsymbol{h}}_t = A \cdot \widetilde{\boldsymbol{h}}_t \quad (17)$$

This operation is a graph convolution process on the feature dimensions, where the final expression of each feature dimension is not only determined by itself but also integrates other highly relevant feature information, forming a structure-aware higher-order feature representation. The enhanced feature representation $\widehat{\boldsymbol{h}}_t$ will be passed as a sequence into the subsequent Transformer-based prediction module for long-sequence global modeling and multi-step order volume prediction.

*3.5 Structure-Aware Self-Attention Mechanism*

To enhance the model's ability to model the internal structure of input features, we introduce a structure-aware self-attention mechanism (SSA) into the sequence modeling backbone network. This mechanism utilizes the feature dependency relationship matrix learned in the first stage to construct a structural bias matrix, guiding the attention mechanism to focus more on feature dimensions with strong dependencies during information interaction, thereby improving the model's ability to perceive and express the internal structure of multimodal fusion representations. The improved model not only inherits the efficiency advantages of Informer in long-sequence prediction but also possesses the capability of collaborative perception across regions.

We use the feature representation enhanced by regional common prompt fusion and structural weighting as input and further introduce time information for embedding. The time encoding function (TE) is applied to obtain the time embedding vector $\mathsf{TimeEmbedding}(t) \in \mathbb{R}^d$, while the position encoding embedding $\mathsf{PE}(\hat{\mathbf{h}}_t) \in \mathbb{R}^d$ is also obtained. The final input representation is:

$$\boldsymbol{e}_t = \hat{\boldsymbol{h}}_t + \mathsf{TE}(t) + \mathsf{PE}\left(\hat{\boldsymbol{h}}_t\right) \quad (18)$$

The final $T$ sequence inputs are formed as follows:

$$\{\boldsymbol{e}_1, \boldsymbol{e}_2, \dots, \boldsymbol{e}_T\} \in \mathbb{R}^{T\times d} \quad (19)$$

However, the standard attention mechanism aggregates information solely based on content relevance, neglecting the structural priors between input features. In our model, these structural relationships are particularly critical, such as the enhancing effect of "holidays" on "order volume" and the moderating effect of "weather" on "passenger count." To enhance the model's ability to perceive input structures, we introduce a structural bias term into the original attention calculation of the encoder. For the standard multi-head self-attention module, its computation formula is:

$$\mathsf{ATT}(\mathbf{Q},\mathbf{K},\mathbf{V}) = \mathsf{SoftMax}\left(\frac{\mathbf{Q}\mathbf{K}^\top}{\sqrt{d}}\right)\mathbf{V} \quad (20)$$

We extend $\mathsf{ATT}(\mathbf{Q},\mathbf{K},\mathbf{V})$ to a structure-enhanced attention mechanism by introducing a structural bias term before the SoftMax operation:

$$\mathsf{ATT}(\mathbf{Q},\mathbf{K},\mathbf{V}) = \mathsf{SoftMax}\left(\frac{\mathbf{Q}\mathbf{K}^\top}{\sqrt{d}} + \mathbf{B}_A\right)\mathbf{V} \quad (21)$$

Where the structural bias term $\mathbf{B}_A \in \mathbb{R}^{d\times d}$ is constructed from the feature relationship matrix $\boldsymbol{A}$:

$$\mathbf{B}_A[i,j] = \log\left(1 + \boldsymbol{A}[i,j]\right) \quad (22)$$

To smooth the original structural relationship values and prevent overly large biases from causing gradient instability, a logarithmic transformation is applied. By embedding structural priors between features into the attention weight computation, we effectively enhance the model's ability to understand the internal semantic structure of the inputs. The structure-enhanced attention mechanism not only strengthens interaction modeling between multimodal information but also improves feature selectivity and temporal generalization in sequence prediction tasks, further enhancing the model’s performance in complex traffic time series forecasting scenarios.

*3.6 Optimization Objective*

We use Mean Squared Error (MSE) as the metric for prediction error, and combine it with the alignment loss of the structural prompt vectors from the first stage to form a joint optimization objective:

$$\mathcal{L} = \sum_{t=1}^{T'} \left(\hat{y}_t - y_t\right)^2 + \lambda\left\|\boldsymbol{p}_s - \boldsymbol{p}_t\right\|^2 \quad (23)$$

Where $\hat{y}_t$ denotes the predicted order volume at time step $t$, and $y_t$ is the corresponding ground truth. $\boldsymbol{p}_s$ and $\boldsymbol{p}_t$ represent the prompt vectors for the structured modality and the textual modality, respectively. $\lambda$ is a balancing coefficient. This loss function not only optimizes prediction accuracy but also guides the semantic alignment of prompt vectors across modalities, effectively enhancing the stability of the fused representations.

## IV. Experimental Setup

### *4.1 Datasets*

We conducted experiments on three datasets: the taxi public dataset from NYC Taxi & Limousine Commission, TLC (October and November, 2024), a private dataset in Chengdu from DiDi ( November, 2016), and a private Bike-sharing dataset from The City of New York's bicycling data (October and November, 2024).

### *4.2 Evaluation Metrics*

To comprehensively assess the performance of the proposed prediction model in real-world. They're widely regarded for their interpretability and general applicability in real-world tasks. Scenarios, we employ three widely used regression metrics for quantitative evaluation: Mean Absolute Error (MAE), Root Mean Square Error (RMSE), and Mean Absolute Percentage Error (MAPE). These metrics quantify the discrepancies between predicted and actual values from various perspectives and are highly regarded for their interpretability and broad applicability..

MAE measures the average absolute difference between the predicted values and the actual values, and is defined as follows:

$$\mathrm{MAE} = \frac{1}{T}\sum_{t=1}^{T} \left|\widehat{y}_t - y_t\right| \tag{24}$$

Where $\widehat{y}_t$ denotes the predicted value at time step $t$, $y_t$ is the corresponding ground truth, and $T$ represents the total length of the prediction sequence. MAE reflects the average deviation between the predicted results and the actual values. Its unit is consistent with that of the target value, and a smaller MAE indicates lower model error.

RMSE measures the square root of the mean of the squared prediction errors and is defined as follows:

$$\mathrm{RMSE} = \sqrt{\frac{1}{T}\sum_{t=1}^{T} (\widehat{y}_t - y_t)^2} \tag{25}$$

RMSE amplifies the parts of the prediction that deviate significantly from the ground truth, placing greater emphasis on penalizing extreme errors. This helps the model identify and improve its ability to predict peaks and large fluctuations.

MAPE measures the prediction error as a percentage relative to the actual value and is defined as follows:

$$\mathrm{MAPE} = \frac{100\%}{T}\sum_{t=1}^{T} \left|\frac{\widehat{y}_t - y_t}{y_t}\right| \tag{26}$$

MAPE provides a unit-free metric to measure the *relative error* of the predictions, indicating the average percentage deviation of the predicted values from the actual values. However, when the ground truth $y_t$ approaches zero, the error can be significantly amplified or even tend toward infinity, making it unstable in low-value scenarios. Therefore, it is necessary to set a lower bound or filter out outliers when using this metric.

### *4.3 The Comparison Methods*

Eight state-of-the-art algorithms are used as the comparison methods. To comprehensively analyze and benchmark the performance of the proposed method, we provide detailed introductions to the design principles, core modules, and innovations of each baseline algorithm.

MM[41] is a multimodal event-based traffic prediction algorithm specifically designed to address the sudden impacts of diverse large-scale urban events (e.g., exhibitions, concerts, sports games) on surrounding traffic.

STAEformer[42] integrates spatiotemporal adaptive embedding with a standard Transformer architecture to significantly enhance traffic time-series forecasting. STAEformer stands out by outperforming advanced baselines on six real-world traffic datasets purely through smart input embeddings.

STDN[43] achieves traffic flow prediction via a dynamic relational graph learning module, a spatiotemporal embedding module, and a trend-seasonal decomposition module. It can better capture global traffic dynamics and significantly improve accuracy, and has better performance in spatiotemporal modeling and multi-resolution feature fusion.

SSL-STMFormer[44] combines self-supervised learning with spatiotemporal entanglement-aware mechanisms to effectively model complex, dynamic dependencies using a Transformer architecture, and it performs excellently in capturing long-range temporal, spatial, and cross-regional dynamics.

STID[45] is a multivariate time series prediction method that addresses the indistinguishability of samples in time and space, and it outperforms popular models in both accuracy and computational efficiency, validating the idea that even simple architectures can achieve state-of-the-art performance.

DTG-MGCN[46] is tailored for taxi demand forecasting. It leverages POI (point of interest) data with fuzzy sets and adaptive kernel density estimation to define semantically meaningful urban "places" as basic prediction units, enhancing spatial modeling interpretability.

SOUP[47] is a framework for spatiotemporal demand prediction and competitive supply optimization in traffic services. It captures long- and short-term dependencies and incorporates context like weather and holidays to enhance accuracy.

MLRNN[48] is an innovative taxi demand forecasting algorithm designed to address demand heterogeneity across urban regions. MLRNN outperforms several advanced baselines in both prediction accuracy and robustness.

### *4.4 Experimental Results*

**(1)The comparison experiments**

Table 1 presents the prediction performance of the proposed method and various baseline models on the NYC-Taxi dataset. We can see that our model (KGCM) achieves the best performance across all metrics in both time periods, demonstrating significant predictive advantages. While traditional models such as SOUP, STAE, and SSL-STMFormer exhibit notably higher errors, indicating their limited ability to model multimodal structures and temporal features effectively.

Table 1. Performance comparison of different models on the NYC Taxi dataset

| Method | Oct | | | Nov | | |
|---|---|---|---|---|---|---|
| | MAPE(%) | MAE | RMSE | MAPE(%) | MAE | RMSE |
| **KGCM** | **35.62** | **3.75** | **5.14** | **43.52** | **3.86** | **5.21** |
| MM | 35.90 | 3.97 | 5.90 | 43.70 | 3.91 | 5.95 |
| SOUP | 75.89 | 6.60 | 10.01 | 85.93 | 6.52 | 10.18 |
| STID | 42.33 | 4.57 | 7.10 | 82.74 | 5.93 | 9.03 |
| STAE | 63.53 | 5.92 | 9.57 | 61.86 | 5.19 | 8.22 |
| DTG-MGCN | 52.69 | 4.43 | 7.88 | 53.22 | 5.32 | 8.68 |
| STDN | 45.25 | 4.39 | 6.88 | 51.81 | 4.38 | 6.93 |
| SSL-STMFormer | 85.64 | 7.19 | 10.94 | 146.81 | 11.47 | 17.63 |
| MLRNN | 66.81 | 4.74 | 6.43 | 86.56 | 5.62 | 11.34 |

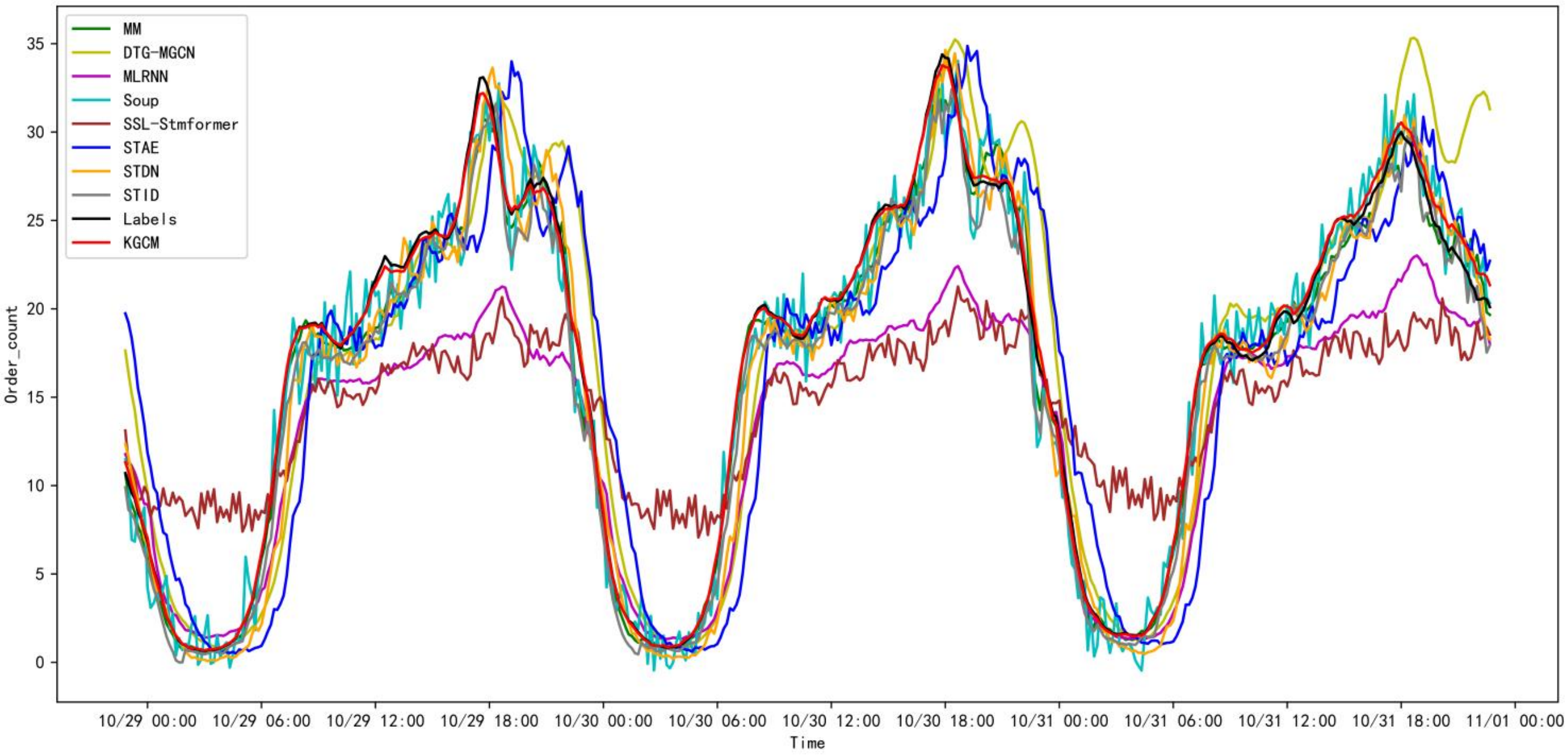

Fig. 3. Visualization Results of nine methods on the NYC Taxi dataset in October

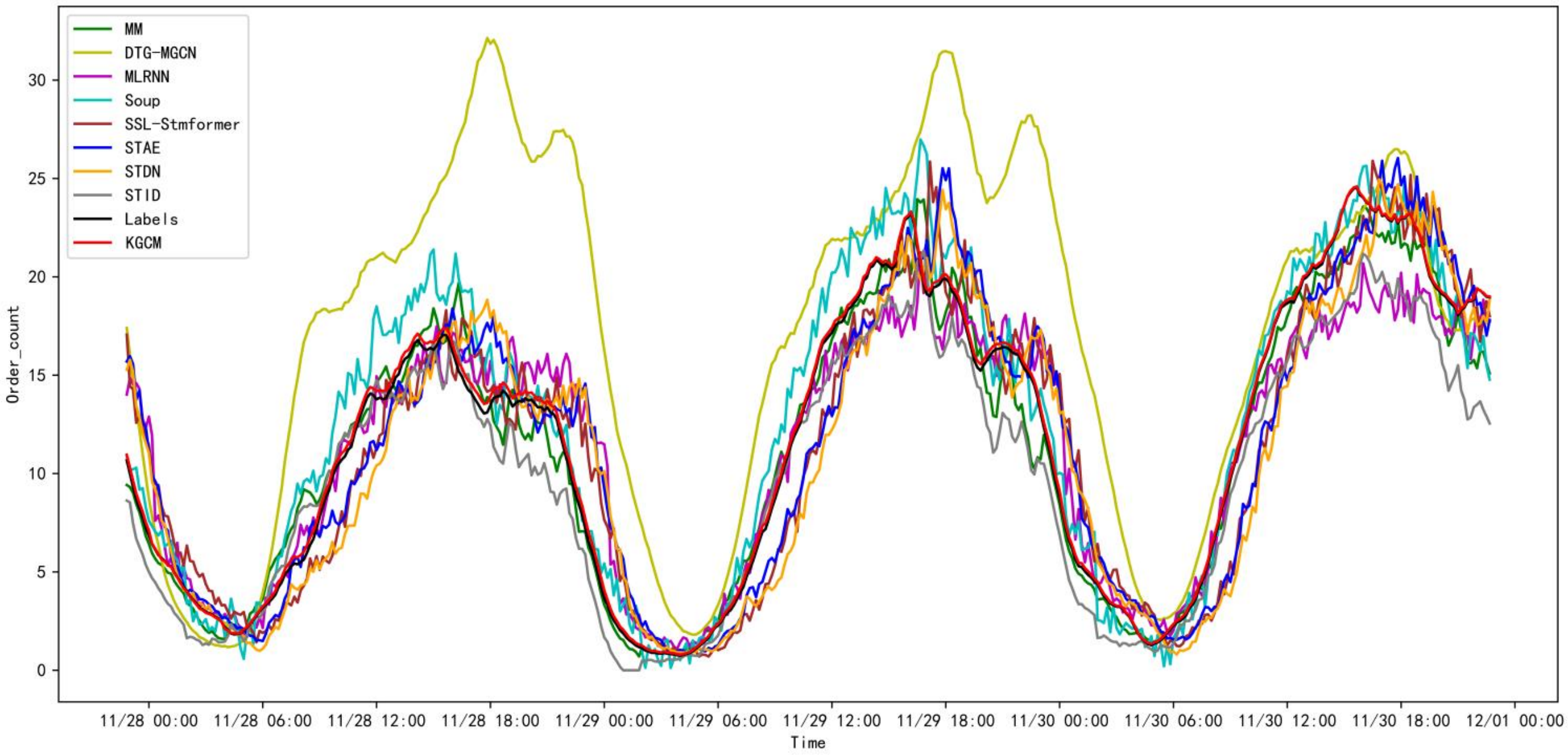

Fig. 4. Visualization Results of nine methods on the NYC Taxi dataset in November

Fig. 3 and Fig. 4 illustrate the performance comparison between the predicted curves of different models and the ground truth labels on the NYC Taxi dataset for October and November, respectively. As shown in these figures, the prediction curve of KGCM closely aligns with the overall trend of the actual labels, particularly in intervals with pronounced periodic fluctuations, where the predicted curve tightly follows the real variations. Moreover, peaks and troughs are accurately captured, demonstrating the model's strong capability in modeling both short-term and long-term temporal dependencies. In contrast, several baseline methods (such as STAE, SOUP, and SSL-STMFormer) exhibit significant deviations at peaks and troughs, especially in the November data, indicating weaker fitting capabilities. STDN and STID provide stable predictions but demonstrate limited responsiveness to large fluctuations. SSL-STMFormer displays certain irregular spikes or abnormal fluctuations in its predictions.

Both KGCM and MM are traffic prediction models that leverage multimodal data fusion. They demonstrate superior predictive performance, outperforming other baseline methods across multiple evaluation metrics. This outcome verifies the significant role that prior traffic knowledge plays in predicting traffic information.

Table 2: Performance comparison of different models on the Chengdu dataset

| Method | Nov | | |
|---|---|---|---|
| | MAPE(%) | MAE | RMSE |
| **KGCM** | **28.89** | **3.96** | **5.23** |
| MM | 30.21 | 4.56 | 6.58 |
| SOUP | 50.84 | 6.35 | 10.54 |
| STID | 32.42 | 4.00 | 6.20 |
| STAE | 43.74 | 5.43 | 8.50 |
| STDN | 42.07 | 5.61 | 8.78 |
| DTG-MGCN | 38.23 | 4.97 | 7.56 |
| SSL-STMFormer | 173.39 | 12.20 | 17.87 |
| MLRNN | 34.29 | 5.15 | 6.29 |

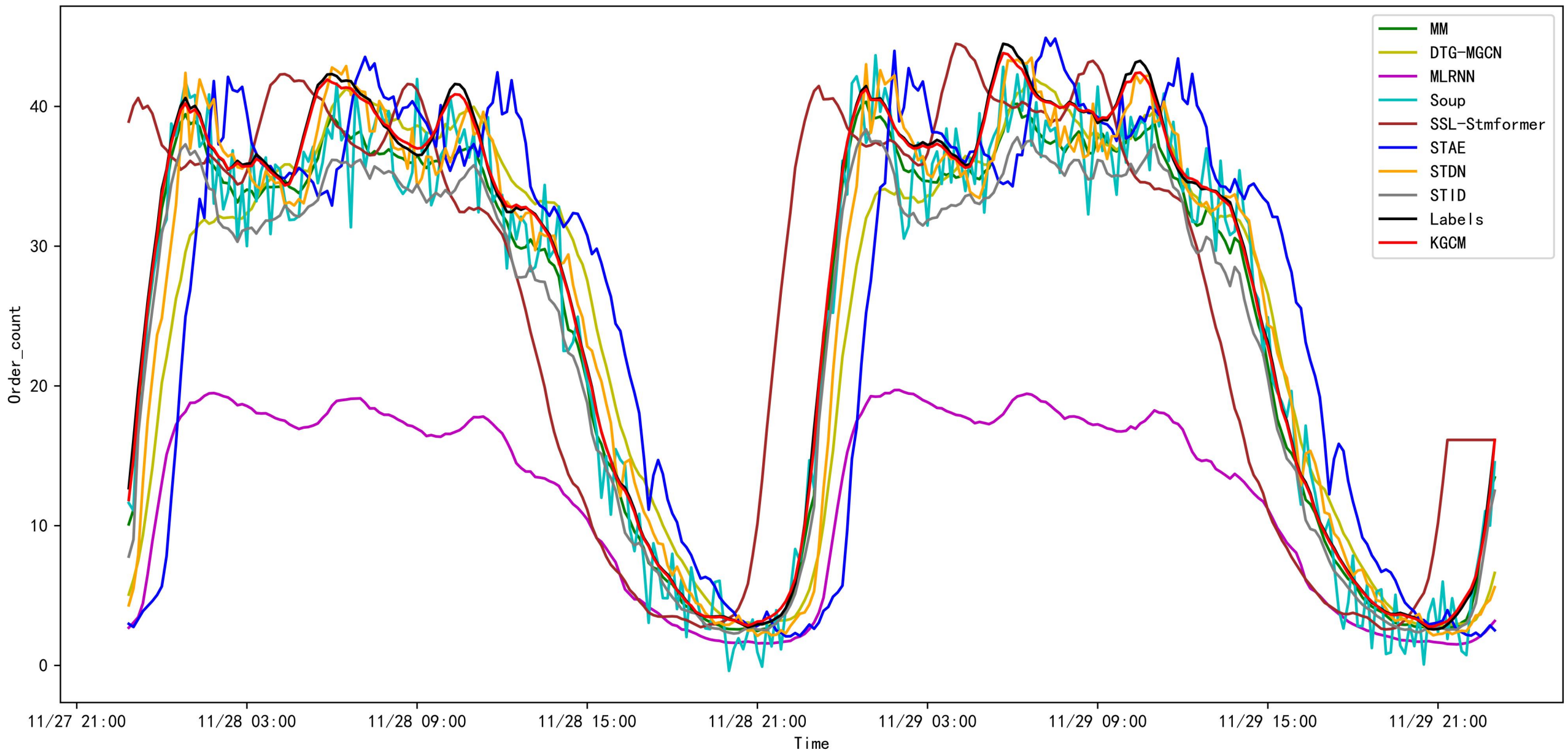

Fig. 5. Visualization Results of nine methods on the Chengdu dataset in November

Table 2 presents the comparison of prediction performance between KGCM and several mainstream models on the Chengdu private dataset. In terms of MAPE, KGCM achieves the best result of 28.89%, indicating its superior capability in handling relative error and better adapting to the scale variations of order volumes across different periods. For MAE, KGCM also outperforms all other models with an MAE value of 3.96. In terms of RMSE, KGCM achieves the best performance with a root mean square error of 5.23. In contrast, SOUP exhibits a much higher RMSE of 10.54, suggesting significant shortcomings in handling outliers and controlling prediction fluctuations.

Fig. 5 presents a visual comparison of the prediction results of nine different methods on the Chengdu dataset (November). It is observed that the results of KGCM (red line) closely align with the ground truth (black line), effectively capturing the fluctuations during morning and evening peak hours as well as the valleys, with outstanding performance in both the position and amplitude of peaks and troughs. In contrast, the methods like STAE (brown line) and SOUP (light blue line) show significant deviations, particularly underperform during low-demand periods, where predictions are either overly high or exhibit abnormal fluctuations. Meanwhile, some methods, such as SSL-STMFormer (purple line) and STID (dark blue line), demonstrate stable overall trends but struggle to accurately fit the peak values. Overall, KGCM demonstrates superior predictive performance in terms of trend alignment, peak detection, and fluctuation responsiveness, validating its effectiveness and robustness in complex traffic forecasting scenarios.

Table 3: Performance comparison of different models on the bike dataset

| Method | Oct | | | Nov | | |
|---|---|---|---|---|---|---|
| | MAPE(%) | MAE | RMSE | MAPE(%) | MAE | RMSE |
| **KGCM** | **32.68** | **3.39** | **4.92** | **53.68** | **2.47** | **3.36** |
| MM | 33.76 | 3.75 | 5.15 | 54.22 | 2.54 | 3.44 |
| SOUP | 91.32 | 7.65 | 10.96 | 89.17 | 4.11 | 6.28 |
| STID | 79.85 | 3.42 | 5.00 | 89.20 | 3.27 | 4.62 |
| STAE | 42.39 | 4.05 | 6.08 | 57.57 | 2.74 | 3.86 |
| STDN | 43.68 | 3.94 | 5.86 | 58.92 | 2.64 | 3.87 |
| DTG-MGCN | 55.47 | 3.89 | 5.21 | 61.23 | 2.74 | 3.88 |
| SSL-STMFormer | 139.89 | 8.25 | 11.29 | 170.48 | 5.56 | 6.95 |
| MLRNN | 75.75 | 3.65 | 3.49 | 72.93 | 3.34 | 3.92 |

Table 3 presents the performance comparison of different models on the bike dataset in October and November. It is evident that KGCM achieves the best results across all metrics, which demonstrates the significant advantages of KGCM over other methods. For the traffic data in October, KGCM attains an MAPE of 32.68%, which is lower than the second-best model of MM (33.76%), and substantially better than traditional models such as STID (79.85%), STDN (43.68%), and SSL-STMFormer (139.89%). Additionally, KGCM achieves the lowest MAE and RMSE values of 3.39 and 4.92, respectively, which indicates superior control over both absolute error and error variance. While for the traffic data in November, KGCM continues to lead in performance with MAPE, MAE, and RMSE values of 53.68%, 2.47, and 3.36, respectively. These results not only demonstrate superior accuracy compared to other methods but also highlight the model's stability across different periods, indicating strong generalization ability and robustness.

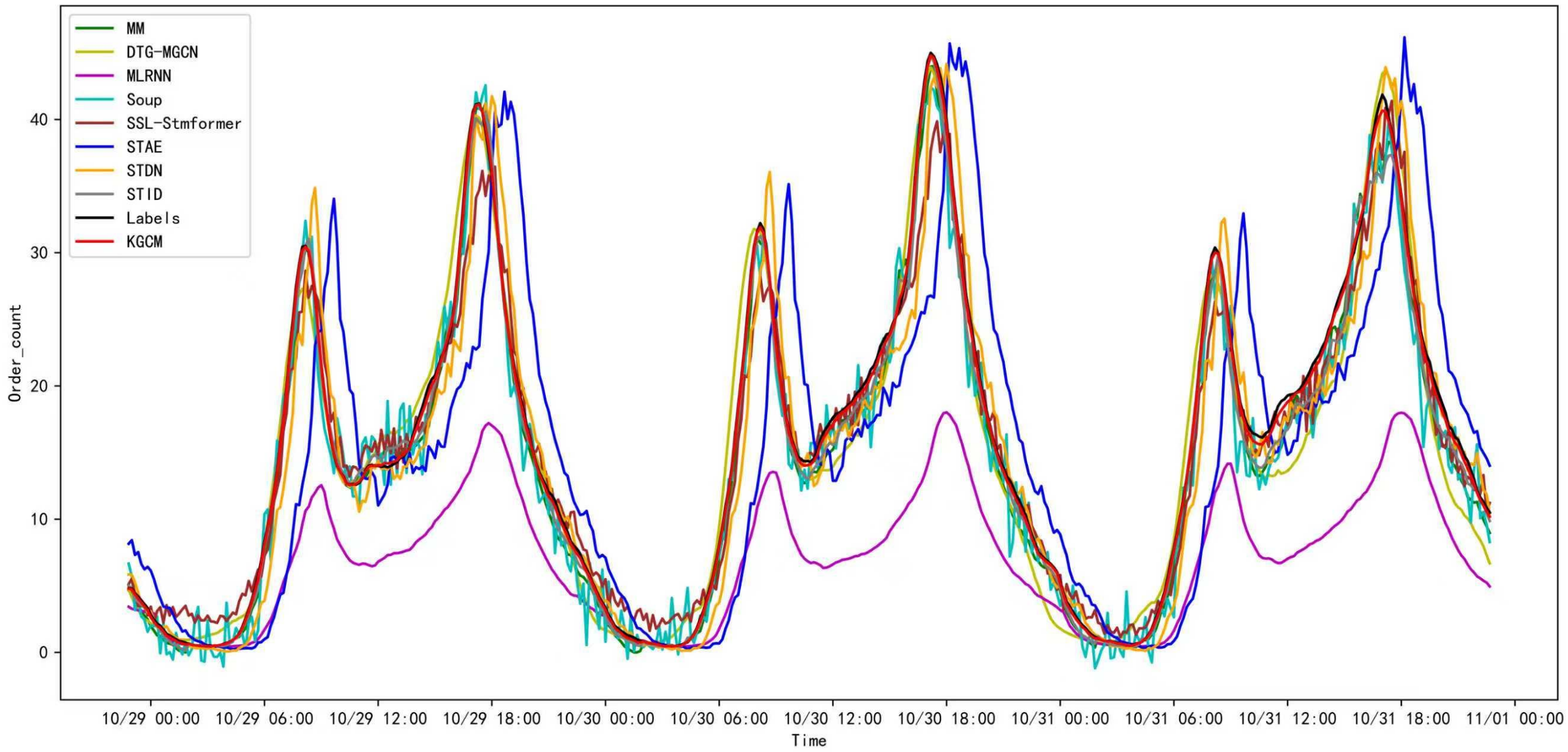

Fig. 6. Visualization Results of nine methods on the Bike dataset in October

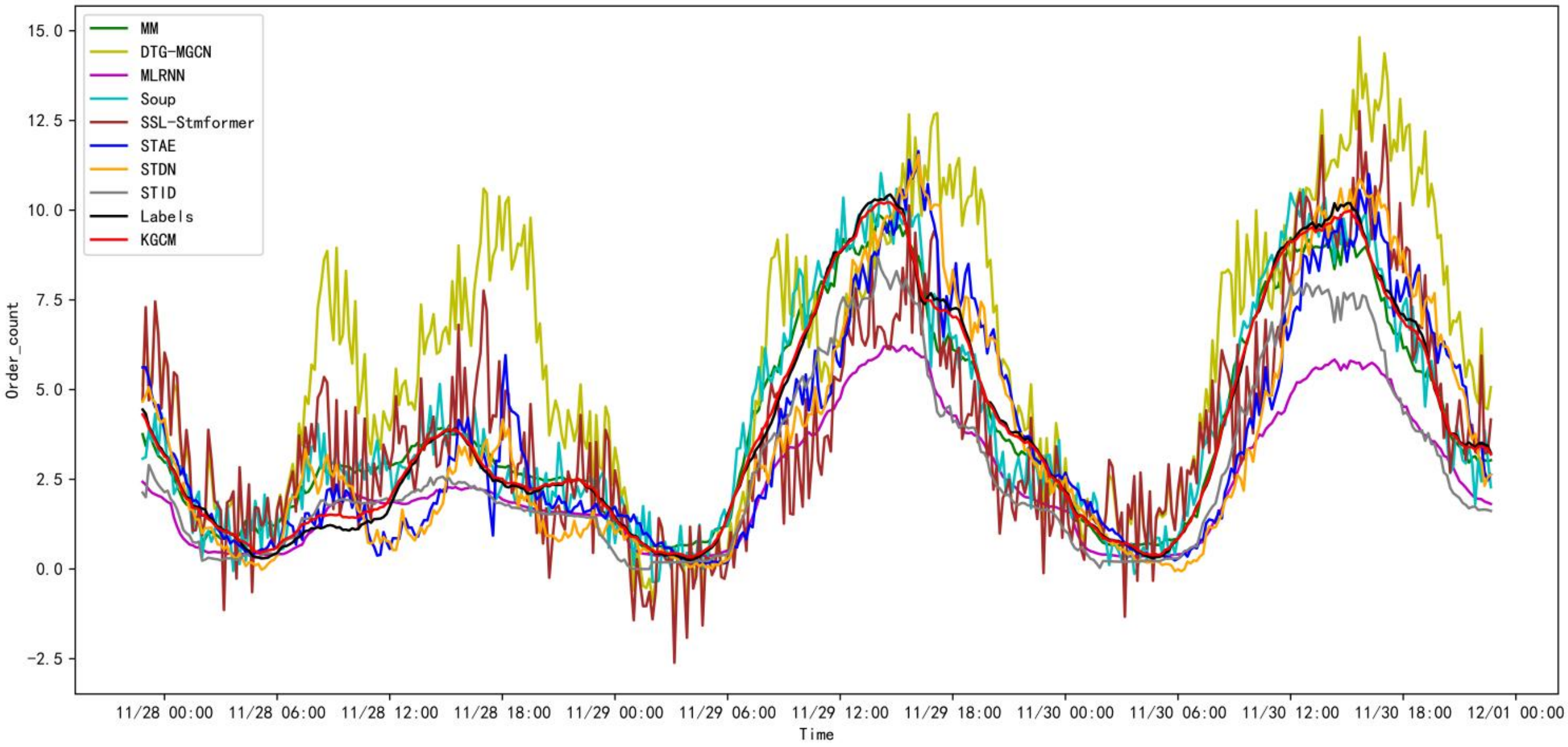

Fig. 7. Visualization Results of nine methods on the Bike dataset in November

Fig. 6 and Fig. 7 present a visual comparison of the predictions from nine different methods on the Bike dataset in October and November, respectively. As shown in Fig. 6, KGCM (red line) closely matches the ground truth (gray line). It accurately fits the peaks and troughs, especially during multiple morning and evening peak periods. In contrast, methods such as SOUP (light blue line) and SSL-STMFormer (purple line) significantly underestimate the peak values, resulting in overly flat prediction curves. STAE (brown line) exhibits considerable fluctuations in certain time intervals, indicating instability in its predictions. Fig. 7 shows the prediction results in November, where KGCM demonstrates the strongest fitting capability through multiple fluctuations, particularly excelling in capturing peak values and recovering from troughs better than other methods. STID (dark blue line) performs abnormally during this month, with excessively high predicted values and unrealistic spikes, while SSL-STMFormer and SOUP display relatively flat trends, lacking sensitivity to actual fluctuations. In summary, KGCM consistently shows superior trend fitting, peak and valley detection, and overall stability across both periods, validating its cross-month robustness and generalization capability on the Bike dataset.

**(2) The ablation experiments**

We further conducted ablation experiments to evaluate the effectiveness of each key component in KGCM. These step-by-step ablation experiments were performed on the three real-world datasets mentioned above, with results shown in Tables 4-6. As more modules were progressively added, the model's performance consistently improved across all evaluation metrics, indicating that each component contributes positively to the overall model. Specifically, the Local Prompt Optimization (LPO) multimodal fusion module had the most significant impact on performance. For example, on the Bike dataset in October, the MAPE value dropped from 41.83% to 32.68% after incorporating LPO. This result confirms the module's role in accurately aligning multimodal features and guiding the model's attention to critical temporal-semantic information. The Adaptive Correlation Matrix-based Feature Weighting (ACMFW) was the second most influential module, with significant improvements in RMSE and MAE demonstrating its importance in eliminating redundant information and enhancing the consistency of feature representations. In addition, the Dynamic Graph Structure Optimization (DGSO) strategy primarily addresses structural relationships between features and shows notable reductions in RMSE and MAE on both the Chengdu and NYC datasets, highlighting its strength in modeling inter-feature structural interactions. The Regional Common Prompt Guidance (RCPG) mechanism, which introduces shared textual information across regions, mainly improves the model's generalization ability. This is most evident in the MAE metric, indicating its contribution to modeling regional consistency. Although the Structure-Aware Self-Attention (SSA) mechanism yields relatively modest numerical gains, it still provides auxiliary benefits in multimodal context understanding, particularly contributing to the stability of RMSE. In summary, the five core modules proposed in KGCM serve distinct yet complementary functions. The ablation results validate the rationality and effectiveness of each component and further demonstrate that KGCM's performance gains arise from the synergistic effect of these components working together.

Table 4: Effectiveness of each key component in KGCM on New York Datasets

| | Components | | | | | | New York Datasets | | |
|---|---|---|---|---|---|---|---|---|---|
| | | | | | | | Oct | | |
| | backbone | SSA | RCPG | DGSO | ACMFW | LPO | MAPE | MAE | RMSE |
| 1 | ✓ | | | | | | 46.76% | 4.79 | 6.28 |
| 2 | ✓ | ✓ | | | | | 45.25%(-1.51%) | 5.48(+0.69) | 5.93(-0.35) |
| 3 | ✓ | ✓ | ✓ | | | | 42.87%(-2.38%) | 5.25(-0.23) | 5.56(-0.37) |
| 4 | ✓ | ✓ | ✓ | ✓ | | | 39.18%(-3.69%) | 4.36(-0.89) | 5.31(-0.25) |
| 5 | ✓ | ✓ | ✓ | ✓ | ✓ | | 37.95%(-1.23%) | 4.14(-0.22) | 5.18(-0.13) |
| 6 | ✓ | ✓ | ✓ | ✓ | ✓ | ✓ | 35.62(-2.33%) | 3.75(-0.39) | 5.14(-0.04) |
| | Components | | | | | | Nov | | |
| | backbone | SSA | RCPG | DGSO | ACMFW | LPO | MAPE | MAE | RMSE |
| 1 | ✓ | | | | | | 61.15% | 5.23 | 6.65 |
| 2 | ✓ | ✓ | | | | | 59.71%(-1.44%) | 4.91(-0.32) | 6.42(-0.23) |
| 3 | ✓ | ✓ | ✓ | | | | 58.33%(-1.38%) | 4.98(+0.07) | 6.32(-0.10) |
| 4 | ✓ | ✓ | ✓ | ✓ | | | 56.75%(-1.58%) | 4.67(-0.31) | 5.98(-0.34) |
| 5 | ✓ | ✓ | ✓ | ✓ | ✓ | | 50.19%(-6.56) | 4.32(-0.35) | 5.62(-0.36) |
| 6 | ✓ | ✓ | ✓ | ✓ | ✓ | ✓ | 43.52%(-6.67) | 3.86(-0.46) | 5.21(-0.41) |

Table 5: Effectiveness of each key component in KGCM on Cheng Dou Datasets

| | Components | | | | | | Cheng Dou Datasets | | |
|---|---|---|---|---|---|---|---|---|---|
| | | | | | | | Nov | | |
| | backbone | SSA | RCPG | DGSO | ACMFW | LPO | MAPE | MAE | RMSE |
| 1 | ✓ | | | | | | 40.16% | 5.12 | 6.89 |
| 2 | ✓ | ✓ | | | | | 36.82%(-3.34%) | 5.26(+0.14) | 6.56(-0.33) |
| 3 | ✓ | ✓ | ✓ | | | | 34.56%(-2.26%) | 5.09(-0.17) | 6.12(-0.44) |
| 4 | ✓ | ✓ | ✓ | ✓ | | | 31.82%(-2.74%) | 4.78(-0.31) | 5.73(-0.39) |
| 5 | ✓ | ✓ | ✓ | ✓ | ✓ | | 31.06%(-0.76%) | 4.69(-0.09) | 5.42(-0.31) |
| 6 | ✓ | ✓ | ✓ | ✓ | ✓ | ✓ | 28.89%(-2.17%) | 3.96(-0.73) | 5.23(-0.19) |

Table 6: Effectiveness of each key component in KGCM on Bike Datasets

| | Components | | | | | | Bike Datasets | | |
|---|---|---|---|---|---|---|---|---|---|
| | | | | | | | Oct | | |
| | backbone | SSA | RCPG | DGSO | ACMFW | LPO | MAPE | MAE | RMSE |
| 1 | ✓ | | | | | | 52.18% | 4.98 | 5.96 |
| 2 | ✓ | ✓ | | | | | 49.86%(-2.32%) | 4.61(-0.37) | 5.73(-0.23) |
| 3 | ✓ | ✓ | ✓ | | | | 46.63%(-3.23%) | 4.42(-0.19) | 5.48(-0.25) |
| 4 | ✓ | ✓ | ✓ | ✓ | | | 46.96%(+0.33%) | 4.12(-0.30) | 5.22(-0.26) |
| 5 | ✓ | ✓ | ✓ | ✓ | ✓ | | 41.83%(-5.13%) | 3.83(-0.29) | 5.03(-0.19) |
| 6 | ✓ | ✓ | ✓ | ✓ | ✓ | ✓ | 32.68%(-9.15%) | 3.39(-0.44) | 4.92(-0.11) |
| | Components | | | | | | Nov | | |
| | backbone | SSA | RCPG | DGSO | ACMFW | LPO | MAPE | MAE | RMSE |
| 1 | ✓ | | | | | | 68.14% | 4.68 | 6.25 |
| 2 | ✓ | ✓ | | | | | 68.57%(+0.43%) | 4.35(-0.33) | 5.89(-0.36) |
| 3 | ✓ | ✓ | ✓ | | | | 66.93%(-1.64%) | 4.10(-0.25) | 5.61(-0.28) |
| 4 | ✓ | ✓ | ✓ | ✓ | | | 61.11%(-5.82%) | 3.90(-0.20) | 5.33(-0.28) |
| 5 | ✓ | ✓ | ✓ | ✓ | ✓ | | 56.32%(-4.79%) | 3.66(-0.24) | 5.04(-0.29) |
| 6 | ✓ | ✓ | ✓ | ✓ | ✓ | ✓ | 53.68%(-2.64%) | 2.47(-1.19) | 3.36(-1.68) |

## V. Conclusion

This paper introduces a knowledge-guided cross-modal feature fusion learning model for traffic demand prediction, which innovatively integrates structured temporal traffic data with textual descriptions of human knowledge. By incorporating local and global adaptive graph structures along with cross-modal feature fusion mechanisms, the model enables efficient coupling and representation of multimodal information. The proposed dynamic graph structure reasoning and updating strategy allows multimodal features to guide the adaptive optimization of model parameters, thereby enhancing prediction accuracy and robustness. Extensive experiments on multiple real-world traffic datasets demonstrate that the proposed model outperforms state-of-the-art traffic prediction methods across all evaluation metrics, particularly in capturing regional and global traffic demand patterns. These results affirm the substantial value of incorporating human knowledge in complex traffic scenarios and underscore the considerable potential of knowledge-guided cross-modal deep learning models in intelligent transportation systems. Looking ahead, we will prioritize enhancing the model's generalization across diverse cities and scenarios, while exploring efficient inference mechanisms to better support real-time demand prediction and decision-making in smart transportation systems.